\documentclass{Interspeech2024}

\interspeechcameraready

\title{Predefined Prototypes for Intra-Class Separation and Disentanglement}

\name[affiliation={1}]{Antonio}{Almudévar}
\name[affiliation={2,3}]{Théo}{Mariotte}
\name[affiliation={1}]{Alfonso}{Ortega}
\name[affiliation={2}]{Marie}{Tahon}
\name[affiliation={1}]{Luis}{Vicente}
\name[affiliation={1}]{Antonio}{Miguel}
\name[affiliation={1}]{Eduardo}{Lleida}

\address{
  $^1$ViVoLab, Aragón Institute for Engineering Research (I3A), University of Zaragoza, Spain\\
  $^2$LIUM, Le Mans University, Le Mans France\\
  $^3$LTCI, Télécom Paris, Institut Polytechnique de Paris, Palaiseau, France
}
\email{almudevar@unizar.es}

\keywords{prototypical learning, predefined prototypes, intra-class separability, disentanglement, explainability}

\begin{document}

\maketitle

% the abstract here must exactly match the abstract entered into the paper submission system
\begin{abstract}
    Prototypical Learning is based on the idea that there is a point (which we call prototype) around which the embeddings of a class are clustered. It has shown promising results in scenarios with little labeled data or to design explainable models. Typically, prototypes are either defined as the average of the embeddings of a class or are designed to be trainable. In this work, we propose to predefine prototypes following human-specified criteria, which simplify the training pipeline and brings different advantages. Specifically, in this work we explore two of these advantages: increasing the inter-class separability of embeddings and disentangling embeddings with respect to different variance factors, which can translate into the possibility of having explainable predictions. Finally, we propose different experiments that help to understand our proposal and demonstrate empirically the mentioned advantages.
\end{abstract}

\section{Introduction} \label{sec:intro}
Prototype theory is a concept from cognitive science and linguistics that suggests that humans categorize objects and ideas by comparing them to a mental representation of a prototype of each category or class \cite{rosch1973natural}. These prototypes encapsulate the most central and representative features of each class. On the other hand, many machine learning systems base their operation on extracting representations \cite{bengio2013representation, achille2018emergence} (from now on we will refer to them interchangeably as embeddings), which are typically vectors that somehow encode the most representative features of the inputs. An approach that has gained popularity due to its good performance in a multitude of applications consists of introducing the idea of prototyping in machine learning systems \cite{klys2018learning}. These systems allow, in addition to extracting a representation of each input, to obtain a representation with the most representative features of the inputs of a class, which is called prototype. Moreover, these systems are typically trained so that all representations of the same class are close together in space. Therefore, as a result of these approximations, we end up having several sets of representations, each one in an area of space and corresponding to a class. Each of these sets of representations is around the prototype of that class. However, there are two properties that in general are not de facto fulfilled in this type of systems and that may be desirable for different tasks and applications. These desirable properties are:
\begin{itemize}[]
  \item The embeddings of different classes are separated in space. This better use of all available space and usually results in a better performance in tasks such as classification, anomaly detection, object detection or biometric recognition \cite{wen2016discriminative, lin2017focal, ranasinghe2021orthogonal}.
  \item It is possible to associate some concrete dimensions of these representations with concrete human-understandable features so that a change of a feature produces changes in only a few dimensions of the space. This is has some advantages such as
  \begin{enumerate*}[label=(\roman*)]
      \item having more control over data creation in generative models \cite{higgins2017beta}, or
      \item providing the ability to explain and interpret model predictions \cite{zhu2021and}.
    \end{enumerate*}
\end{itemize}

In this paper we propose a modification on the prototypical systems that preserves their default advantages and, in addition, allows solving the two problems presented. This modification consists in having the human predefine the prototypes before the training of the system, so that one of the objectives is that all the representations of a class are in an area of the space that has been decided with a human criterion and therefore can be explained. Typically, prototypes are trainable or simply computed as the average of the representations of a class and, to the best of our knowledge, the approach of imposing concrete conditions on them prior to training is an approach not explored in the literature. This may have several advantages, but in this paper we focus on its ability to achieve the two properties explained in the previous paragraph and whose are immediate to understand. First, if we predefine the prototypes before training, we can impose that they are far apart, by defining them as an orthogonal set of vectors, for example. On the other hand, we can define the prototypes so that some of their dimensions correspond to concrete human-interpretable factors.

To demonstrate with examples the two previous points, we propose two types of experiments demonstrating with them the correct operation of our proposal. In the first one, we solve a typical audio classification problem where we impose that representations of different classes are orthogonal to each other, showing that an implication of this is a better accuracy. On the other hand, we solve an emotion classification task where concrete dimensions of the representations correspond to acoustic parameters, allowing an explanation of how these parameters relate to the different emotions to be classified. More details on these experiments can be found at
\ifinterspeechfinal
     \url{https://github.com/antonioalmudevar/predefined_prototypes}
\else
    \url{https://github.com/not/available/in/the/version/for/review}.
\fi

\section{Related Work} \label{sec:related_work}
\textbf{Prototypical Learning.} 
%It is the set of methods that obtain a representation that allows to describe a class or set of elements and that receives the name of prototype. 
Its use was initially proposed for few-shot classification and class prototypes were calculated as the average of the few embeddings available for each class. During training, it is imposed that all embeddings are close to the prototype of their class, which implies that they are all close in space. Subsequently, different applications and variations of the prototypical systems have been proposed. For example, they have been used for unsupervised domain adaptation \cite{pan2019transferrable} or to build explainable systems \cite{zinemanas2021interpretable, mariotte2024explainable}.

\vspace{1mm}\noindent\textbf{Intra-Class Separation.} 
Due to the reasons explained in section\ref{sec:intro}, different loss functions have been proposed that aim to separate embeddings of different classes in space. Among these loss functions, the following stand out: Center Loss \cite{wen2016discriminative}, Focal Loss \cite{lin2017focal}, Orthogonal Projection Loss \cite{ranasinghe2021orthogonal} or that of the Variational Classifier \cite{almudevar23_interspeech}. The proposal of the present work allows defining the prototypes in such a way that they are far from each other and, consequently, the embeddings of the different classes are also far from each other.

\vspace{1mm}\noindent\textbf{Disentanglement.} Although there is no consensus on its meaning, what is widely accepted is the intuition that it is the separation of the representations in the different variation factors of the data \cite{bengio2013representation, higgins2018towards}. That is, changing a variation factor in the data should change only a part of the representation. 
%A multitude of papers have proposed methods to achieve disentanglement in an unsupervised \cite{higgins2017beta, burgess2018understanding, kim2018disentangling} fashion, i.e., without explicitly telling the model the variance factors over which it should disentangle its representations. 
In \cite{locatello2019challenging} it is shown that it is not possible to achieve unsupervised disentanglement without inductive biases in the data or in the model. Therefore, multiple works have also been proposed focused on supervised disentanglement \cite{locatello2020weakly, estermann2023dava, almudévar2024unsupervised}.

\section{Proposed Method} \label{sec:proposed_method}
%As we have explained in the previous section, many systems have been proposed based on the idea of prototypes, which differ from each other in aspects such as the number of prototypes, the way in which these prototypes are defined or the way in which the final predictions of the system are calculated from these prototypes. In this section we explain our proposal, which has as a common aspect with all the previous ones, the fact that there are a series of embeddings that are representative of the rest and that are what we call prototypes. 

\subsection{General System Description}
Let $\mathcal{D}=\{\{x^{(i)},y^{(i)},\alpha^{(i)}\}\}_{i=1}^n$ be a dataset where $x^{(i)} \in \mathbb{R}^p$ are each of the inputs, $y^{(i)}\in \mathbb{R}^C$ is the vector containing the class to which $x^{(i)}$ belongs and $\alpha^{(i)} \in \mathcal{A}$ is an  abstract containing the factors of interest over $x^{(i)}$. We note that $\alpha^{(i)}$ can be extracted by an additional system and are the factors over which we intend to disentangle our representations.
Deep Learning classifier systems are usually divided into two parts: $F_\theta: \mathbb{R}^p \rightarrow \mathbb{R}^k$, which we call embeddings extractor, and $G_\phi: \mathbb{R}^k \rightarrow \mathbb{R}^C$, which is the classifier network and is typically a linear layer plus a Softmax. On the other hand, in our system, we have $P: \mathbb{R}^C \times \mathcal{A} \rightarrow \mathbb{R}^k$, which we call the prototype extractor. We note that $P$ does not depend on any trainable parameter, that is, it is not modified throughout the training, which is the main novelty of our work.
From these three components we can compute embeddings as $z^{(i)}=F_\theta(x^{(i)})$, predictions as $\tilde{y}^{(i)}=G_\phi(z^{(i)})$ and prototypes as $p^{(i)}=P(y^{(i)}, \alpha^{(i)})$. The two objectives of this system are:
\begin{enumerate*}[label=(\roman*)]
  \item that $y^{(i)}$ and $\tilde{y}^{(i)}$ are similar to have a good classification performance, and
  \item that $z^{(i)}$ and $p^{(i)}$ are similar for the embedding extractor $F_{\theta}$ to behave similarly to $P$.
\end{enumerate*}
This results in the following loss function:
\begin{equation} \label{eq:loss}
    \mathcal{L} = CE\left(y^{(i)}, \tilde{y}^{(i)}\right) + \lambda_p ||z^{(i)} - p^{(i)}||_2^2
\end{equation}
where $CE$ is the cross-entropy loss and $\lambda_p$ is a hyperparameter that we set to $1/k$ in the experiments. In algorithm \ref{alg:training} we summarize the training procedure.

\begin{algorithm}
\caption{Training Algorithm for the Predefined Prototypes System}
\label{alg:training}
\begin{algorithmic}
\Require Dataset $\mathcal{D}$, Prototypes Extractor $P$, $\lambda_p$
\State $\bm{\theta}$, $\bm{\phi} \gets$ Initialize parameters
\Repeat
\State $\mathcal{D}^N \gets \{(x^{(i)}, y^{(i)}, \alpha^{(i)}\}\}_{i=1}^N$ (Minibatch from $\mathcal{D}$)
\For{$i=1$ to $N$}
    \State $z^{(i)} \gets F_\theta(x^{(i)})$
    \State $\tilde{y}^{(i)} \gets G_\phi(z^{(i)})$
    \State $p^{(i)} \gets P(y^{(i)}, \alpha^{(i)})$
\EndFor
\State $\mathcal{L} \gets \frac{1}{N} \sum_{i=1}^N CE\left(y^{(i)}, \tilde{y}^{(i)} \right) + \lambda_p ||z^{(i)} - p^{(i)}||_2^2$
\State $\bm{\theta}, \bm{\phi} \gets$ Update using gradients of $\mathcal{L}$
\Until{convergence of $\bm{\theta}$ and $\bm{\phi}$}
\end{algorithmic}
\end{algorithm}

Finally, we propose that $P$ is a multilinear map, i.e.:
\begin{itemize}
    \item $P(ay^{(i_a)}+by^{(i_b)}, \alpha^{(i)})=aP(y^{(i_a)}, \alpha^{(i)})+bP(y^{(i_b)}, \alpha^{(i)})$, which allows working with soft labels (which implies, among other things, the ability to use regularization techniques such as mixup \cite{zhang2017mixup}, widely used in audio classification).
    \item $P(y^{(i)}, a\alpha^{(i_a)}+b\alpha^{(i_b)})=aP(y^{(i)}, \alpha^{(i_a)})+bP(y^{(i)}, \alpha^{(i_b)})$, which provides ease in managing continuous variation factors.
\end{itemize}

\subsection{Examples of Prototypes Extractors}
The main novelty of our system with respect to others is the proposal that $P$ is not modified during training and can also be defined by the human. This, despite being an unexplored idea to the best of our knowledge, may have a large number of applications. Specifically, in this paper we give two formulas that allow us to solve two very popular problems. These are:
\begin{enumerate*}[label=(\roman*)]
  \item maximize the distance between embeddings of different types and
  \item disentangle subsets of embeddings variables with respect to specified factors of variation.
\end{enumerate*}
In the following, we give the formulas we propose to define $P$ that allow to solve these problems.

\subsubsection{Maximize Intra-Class Separability} \label{subsubsec:intraclass}
In this first example, we have that $P(y^{(i)},\alpha^{(i)})=P(y^{(i)}): \mathbb{R}^C \rightarrow \mathbb{R}^k$, since our only goal is to separate embeddings of different classes independently of the input factors. Subsequently, we define (using Singular Value Decomposition, for example) an orthogonal basis $V=\{v^{(j)}\}_{j=1}^C$ such that $v^{(j)} \in \mathbb{R}^k$, so that $P(j)=v^{(j)}$. In this way, the prototypes of the different classes are orthogonal to each other and, consequently, the embeddings of elements of different classes will tend to be quasi-orthogonal, thus being far apart in space.
Although typically the dimension of the embedding is higher than the number of classes, i.e., $k \geq C$, there may be scenarios in which this is not the case, which would prevent one from being able to define an orthogonal $V$ basis as described. In the case where $k < C$, we propose to define an orthogonal basis $W=\{w^{(j)}\}_{j=1}^C$ such that $w^{(j)} \in \mathbb{R}^C$, subsequently define a Johnson Lindenstrauss Transform (JLT) $T: \mathbb{R}^C \rightarrow \mathbb{R}^k$, so that finally $v^{(j)}=T(w^{(j)}) \in \mathbb{R}^k$ for $j=1,2,\dots,C$. JLTs are nearly distance-preserving transformations, so embeddings of different classes will end up being maximally separated in space \cite{johnson1986extensions}. Fast algorithms exist to compute these JLTs \cite{ailon2006approximate}.

\subsubsection{Embeddings Disentanglement with respect to Features} \label{subsec:emd_disentanglement}
In this second example we define our prototype extractor as $P(y^{(i)}, \alpha^{(i)})=P_\alpha(\alpha^{(i)}): \mathcal{A} \rightarrow \mathbb{R}^k$, i.e., it depends on the factors but not on the labels. The way in which this $P_\alpha(\alpha^{(i)})$ is defined is highly dependent on the scenario and application. However, one way we have found to obtain a good classification performance and disentangled embeddings is to set $P_\alpha(\alpha^{(i)}) = \left( P'_\alpha(\alpha^{(i)}), \bm{0} \right)$, where $(\cdot ,\cdot )$ means concatenation, $P'_\alpha(\alpha^{(i)}): \mathcal{A} \rightarrow \mathbb{R}^{k_f}$ and $\bm{0}$ is the all-zero vector of length $k-k_f$. Thus, the goal is that we can clearly separate our embeddings into two parts:
\begin{enumerate*}[label=(\roman*)]
  \item one that depends on the factors and is disentangled with respect to them, and
  \item one whose value does not depend directly on the factors and whose elements are close to zero.
\end{enumerate*}
The purpose of this second part of the embedding is to give the model a greater degree of freedom in organizing its embeddings and to capture all the factors of variation that influence the predictions $\tilde{y}^{(i)}$ but are not captured in $\alpha^{(i)}$. Without this second part (or, equivalently, if $k=k_f$), predictions would be made from only $\alpha^{(i)}$, which would generally result in lower accuracy, since there are more factors that can affect $\tilde{y}^{(i)}$ and are not in $\alpha^{(i)}$. In fact, by comparing the two parts of the embeddings, we can obtain information the proportion the predictions determined by the known factors $\alpha^{(i)}$ and by unknown factors. 
Although one might think that defining the second part of the prototypes as $\bm{0}$ might lead to its elements tending to always be worth 0, this is not necessarily the case. As is the case in the Variational Autoencoder \cite{kingma2013auto}, the loss function to be optimized defined in the equation \eqref{eq:loss}, is defined by two terms. The fact that cross-entropy must be minimized will lead our embeddings to differ across classes if this helps to minimize cross-entropy. If, on the other hand, our predictions can be made very accurately only from $\alpha^{(i)}$ (which is highly unlikely), the elements of the second part of the embeddings will tend to take values close to 0, which is positive for two reasons:
\begin{enumerate*}[label=(\roman*)]
  \item it indicates that there are no factors other than those in $\alpha^{(i)}$ that affect the predictions, which gives us new knowledge about our dataset; and
  \item it tells us that there are elements of the embeddings that are unnecessary, thus reducing the size of the embeddings and, consequently, the complexity of the model.
\end{enumerate*}
The way to define $P'_\alpha$ is very scenario-dependent, so we do not give a general formula to define it. In section \ref{subsubsec:exp_er} we describe an emotion recognizer in which embeddings are disentangled with respect to a set of acoustic parameters and follow the previous formulation. 

\begin{table*}[t]
\centering
\caption{Accuracy for the different Datasets, Embeddings Extractors and Loss Functions}
\label{tab:acc}
\begin{tabular}{l|cc|cc}
\hline
\multicolumn{1}{c|}{}               & \multicolumn{2}{c|}{ESC-50}                           & \multicolumn{2}{c}{KS2}                       \\
\multicolumn{1}{c|}{}               & AST                       & BEATs                     & AST                       & BEATs                     \\ \hline
Cross-entropy                       & $93.97\pm 0.21$           & $91.05 \pm 0.41$          & $\mathbf{92.05 \pm 0.04}$   & $88.94 \pm 0.13$          \\
Focal Loss                          & $94.40\pm 0.36$           & $91.10 \pm 0.49$          & -                         & -                         \\
OPL                                 & $94.11\pm 0.37$           & $91.50 \pm 0.20$          & -                         & -                         \\ \hline
Predefined Prototypes               & $\mathbf{94.52 \pm 0.02}$   & $\mathbf{91.72 \pm 0.30}$   & $91.45 \pm 0.06$          & $\mathbf{89.42 \pm 0.09}$    \\ \hline
\end{tabular}
\end{table*}

\section{Experiments} \label{sec:experiments}

\subsection{Datasets}
\noindent\textbf{Environmental Sound Classification (ESC-50)} \cite{piczak2015esc} includes 2,000 ambient sound recordings categorized into 5 classes, each lasting 5 seconds. Throughout our experiments, we adhere to the standard 5-fold cross-validation approach.

\noindent\textbf{Speech Commands V2 (KS2)} \cite{warden2018speech} comprises 105,829 one-second clips of spoken keywords, each annotated with one of 35 word classes. Officially, it is segmented into 84,843 training clips, 9,981 test clips, and 11,005 validation clips.

\noindent\textbf{IEMOCAP (ER)} \cite{busso2008iemocap} consists of approximately 12 hours of speech showcasing four distinct emotions. Our evaluation uses the standard 5-fold cross-validation method proposed in \cite{yang2021superb}.

\subsection{Embeddings Extractors}
\noindent\textbf{ECAPA-TDNN} \cite{desplanques2020ecapa} integrates attention mechanisms and parallel processing for efficient sequential data analysis, particularly suited for tasks like speech recognition.

\noindent\textbf{Audio Spectrogram Transformer (AST)} \cite{gong2021ast} renowned as the pioneer in using Transformer architectures for audio, has set a benchmark due to its exceptional performance. We initialize the model with pre-trained weights from Imagenet \cite{deng2009imagenet} and Audioset \cite{gemmeke2017audio} and fine-tune it for each specific scenario.

\noindent\textbf{BEATs} \cite{chen2022beats} is an audio pre-training framework for learning representations from Audio Transformers, in which an acoustic tokenizer and a self-supervised audio model are optimized. We also start from the pre-trained model with Audioset.

\subsection{Results}
In all the next experiments, we use a sampling frequency of 16kHz. The inputs to every system are 128 mel-filterbank calculated in 25 ms windows every 10 ms. We normalize the mean and standard deviation to 0 and 0.5, respectively.

\subsubsection{Audio Classification with Intra-Class Separation}
To compare the performance of our proposal with others in the literature, we evaluate them on ESC-50 and KS2 using AST and BEATs as embedding extractors. The hyperparameters used to obtain the inputs and to train each of them are based on \cite{gong2021ast, chen2022beats}.
To evaluate our proposal, we compare it with identical systems where the only differentiating factor is the loss function. Specifically, we compare our loss function with cross-entropy, Focal Loss and OPL. The latter two, like our proposal, aim at increasing the distance between embeddings of different classes. They give excellent performances in different classification scenarios. We note that we have not been able to train them for KS2, since mixup is used, which results in having soft labels, and these two proposals do not allow working with soft labels. Conversely, our proposal does, since the prototype extractor is a multilinear map. In our proposal, we define the prototypes according to the procedure explained in \ref{subsubsec:intraclass}.
We have run all the described experiments three times and we show the mean and standard deviation of the accuracy in table \ref{tab:acc}. We see that in all cases except one, our system outperforms the rest on average. This seems to indicate that it is effective in separating embeddings of different classes and that this translates into better accuracy. 
\vspace{-2mm}
\begin{figure}[ht!]
    \centering
    \includegraphics[width=0.95\columnwidth]{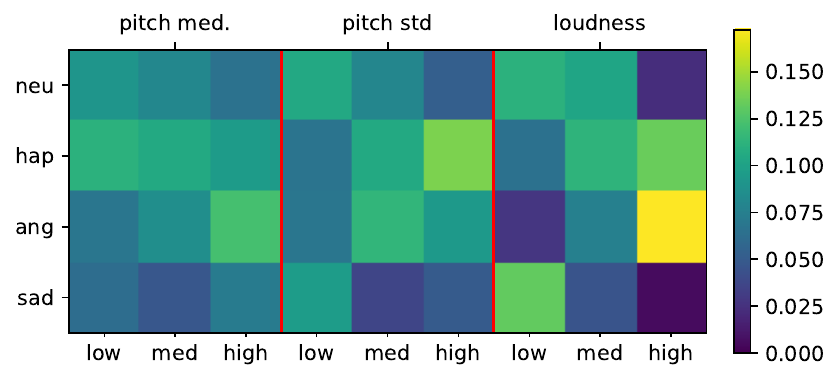}
    \vspace{-2mm}
    \caption{Joint probabilities of each emotion and acoustic parameter (by levels) in the training dataset.}
    \label{fig:joint_prob}
\end{figure}

\vspace{-6mm}
\subsubsection{Disentangled Emotion Recognition} \label{subsubsec:exp_er}
\begin{figure}[ht!]
    \centering
    \begin{subfigure}[b]{0.98\columnwidth}
        \includegraphics[width=\textwidth]{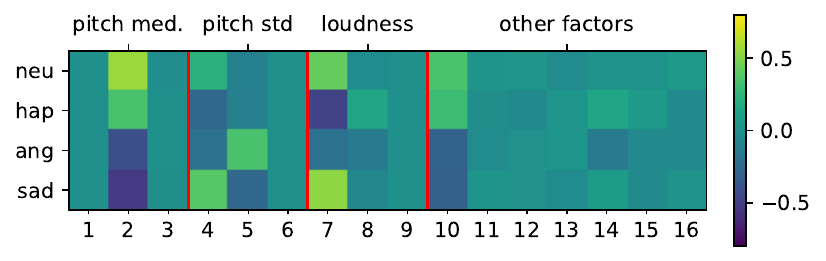}
        \vspace{-5mm}
        \caption{$p_{neu}=0.68$, $p_{hap}=0.15$, $p_{ang}=0.05$, $p_{sad}=0.12$}
        \label{subfig:neu}
        \vspace{2mm}
    \end{subfigure}
    \begin{subfigure}[b]{0.98\columnwidth}
        \includegraphics[width=\textwidth]{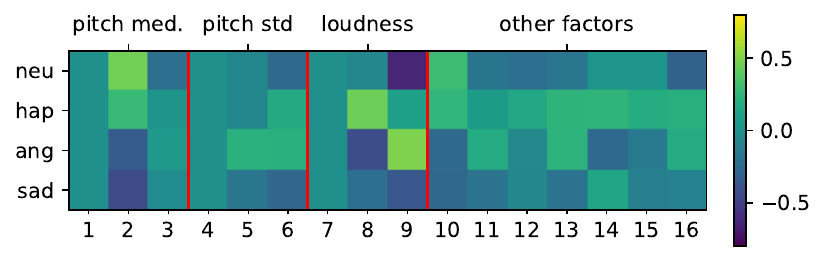}
        \vspace{-5mm}
        \caption{$p_{neu}=0.02$, $p_{hap}=0.87$, $p_{ang}=0.10$, $p_{sad}=0.01$}
        \label{subfig:hap}
        \vspace{2mm}
    \end{subfigure}
    \begin{subfigure}[b]{0.98\columnwidth}
        \includegraphics[width=\textwidth]{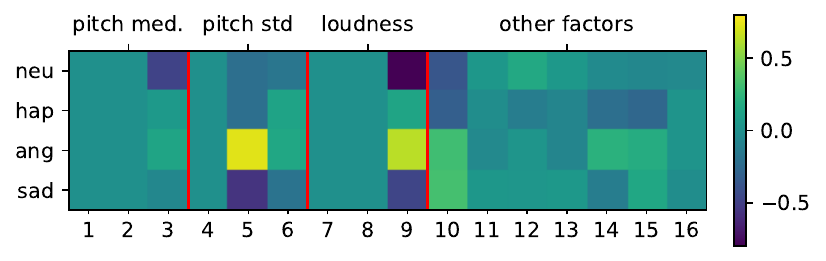}
        \vspace{-5mm}
        \caption{$p_{neu}=0.01$, $p_{hap}=0.04$, $p_{ang}=0.91$, $p_{sad}=0.04$}
        \label{subfig:ang}
        \vspace{2mm}
    \end{subfigure}
    \begin{subfigure}[b]{0.98\columnwidth}
        \includegraphics[width=\textwidth]{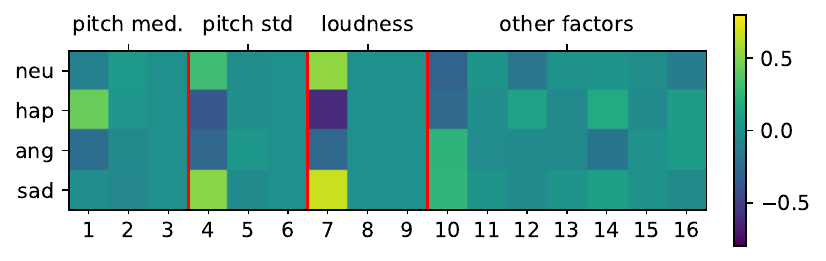}
        \vspace{-5mm}
        \caption{$p_{neu}=0.20$, $p_{hap}=0.10$, $p_{ang}=0.08$, $p_{sad}=0.62$}
        \label{subfig:sad}
        \vspace{2mm}
    \end{subfigure}
    \vspace{-2mm}
    \caption{Relevance matrices $\Gamma^{(i)}$ for different embeddings $z^{(i)}$ and their corresponding predictions $\tilde{y}^{(i)}$}
    \label{fig:embed_import}
\end{figure}

Next we give an illustrative example of how to design a classifier whose embeddings are disentangled with respect to given factors of variation. We propose to use IEMOCAP as dataset to classify the emotions \textit{\{neutral, happy, angry, sad}\} and choose \textit{\{pitch median, pitch std, loudness\}} as factors with respect to which to disentangle our embeddings. We select only female voices, to minimize pitch differences between speakers. The reason for choosing these three acoustic parameters as factors is that they have been found to be related to emotions \cite{murray1993toward, hammerschmidt2007acoustical, sauter2010perceptual}. To extract \textit{pitch median} and \textit{pitch std}, we use \cite{talkin1995robust}; and to obtain \textit{loudness} we follow \cite{recommendation2015itu}.
On the other hand, although $P'_\alpha$ can take continuous values as inputs and can be any type of function, here we discretize the values of the factors to three levels according to their value with respect to the $1/3$ and $2/3$ quantiles calculated in the training set. This facilitates interpretability so that, from now on, we will say that each factor is either \textit{low}, \textit{medium} or \textit{high}. The joint probability of each emotion and level of these factors is shown in Figure \ref{fig:joint_prob}.
Subsequently, we define a coding function $C(\alpha^{(i)}_k) $ as:
\begin{equation}
    \label{eq:coder}
    C(\alpha^{(i)}_k) = \left\{
    \begin{array}{ll}
         (1,0,0)    & \text{if} \, \alpha^{(i)}_k \ \text{is} \ \textit{low} \\
         (0,1,0)    & \text{if} \, \alpha^{(i)}_k \ \text{is} \ \textit{medium}  \\
         (0,0,1)    & \text{if} \, \alpha^{(i)}_k \ \text{is} \ \textit{high} 
    \end{array}
     \right.
\end{equation}
where $\alpha^{(i)}_1$, $\alpha^{(i)}_2$ and $\alpha^{(i)}_3$ are the \textit{pitch median}, \textit{pitch std} and \textit{loudness} of the input $x^{(i)}$, respectively and $\alpha^{(i)} = \left\{\alpha^{(i)}_1,  \alpha^{(i)}_2, \alpha^{(i)}_3\right\}$. Finally, we define $P'_\alpha(\alpha^{(i)}) = \left(C\left(\alpha^{(i)}_1\right), C\left(\alpha^{(i)}_2\right), C\left(\alpha^{(i)}_3\right)\right)$. Thus, an audio with \textit{low pitch median}, \textit{medium pitch std} and \textit{high loudness} would correspond to a prototype $\left( (1,0,0,0,1,0,0,0,1), \bm{0} \right)$. As embedding extractor $F_\theta$ we use an ECAPA-TDNN with embedding size $k=16$ (thus, length of $\bm{0}$ is 7). As classification net  $G_\phi$ we use a linear layer without bias, i.e. $G_\phi \equiv W_\phi = (w^\phi_{jc}) \in \mathbb{R}^{k \times C}$ and the prediction is calculated as $\tilde{y}^{(i)} = W_\phi^T \cdot z^{(i)}$. Equivalently, we define the relevance matrix as:
\begin{equation}
    \Gamma^{(i)} = (\gamma^{(i)}_{jc}) \ \text{such that} \ \gamma^{(i)}_{jc} = w^\phi_{jc} \cdot z^{(i)}_j
\end{equation}
where $\gamma^{(i)}_{jc}$ contains information on how the $j$ component of embedding $z^{(i)}$ influences the probability of class $c$.  Next we analyze these relevance matrices of different embeddings shown in Figure \ref{fig:embed_import} to illustrate this better. 
\begin{itemize}
    \item In Figure \ref{subfig:neu} we see a case where neutral emotion has been predicted with a probability of 0.68. We can see that having low \textit{pitch std} and \textit{loudness} added probabilities to both \textit{sad} and \textit{neutral}, but the fact that the \textit{pitch median} is medium makes the model to decrease the probability of \textit{sad}. As we can see in Figure \ref{fig:joint_prob}, it is uncommon to have medium \textit{pitch median} given the emotion \textit{sad}. 
    \item In Figure \ref{subfig:hap} we can suspect that the \textit{loudness} is between \textit{medium} and \textit{high}, since the activated part of \textit{high} adds probability to \textit{angry}, but the activated part of \textit{medium} subtracts to \textit{angry} and adds to \textit{happy}, which is the emotion with the highest probability. As we can see, the probability of having a \textit{high loudness} when the emotion is \textit{angry} is very high. 
    \item In Figure \ref{subfig:ang} we see clearly that having a \textit{high loudness} and \textit{medium pitch std} has a great influence in predicting \textit{angry}. 
    
    \item In Figure \ref{subfig:sad} we can see that having \textit{low pitch std} and \textit{low loudness} increases the probability of \textit{neutral} and \textit{sad}. However, component 10 has a great influence on ultimately increasing the probability of \textit{sad}. 
\end{itemize}
It is interesting to analyze component 10, since it seems to have similar behavior for \textit{neutral} and \textit{happy} and for \textit{angry} and \textit{sad} in all figures. This raises the suspicion that this component holds a factor related to the positivity or negativity of the emotion, but which we have not tried to disentangle explicitly. This is an example of the importance of giving freedom to a part of the embeddings. This part ends up learning factors that we have not explicitly tried to disentangle but seem to have influence on predictions and interpretable meaning. In addition, we see how the components associated with \textit{other factors} are far from being 0 in most cases, as explained in section \ref{subsec:emd_disentanglement}.
Finally, we mention that the accuracy of the described system is 55.53 and that of one with the same characteristics but with cross-entropy loss 54.86, i.e. we obtain a similar performance in both cases.

\section{Conclusions}
In this paper, we have proposed the possibility of defining the prototypes of a system before training and with a human criterion. We have argued some of the advantages and applications that this may have. Specifically, we have argued how this idea can be used to 
\begin{enumerate*}[label=(\roman*)]
    \item increase the distance between embeddings of different classes and, consequently, improve the accuracy of a classifier; and
    \item disentangle embeddings with respect to given variation factors, being able to have more control over them and providing the possibility to explain predictions.
\end{enumerate*}
However, the ultimate goal of the paper is to give the idea that it may be convenient in some cases to let the human define the prototypes instead of letting the system learn them by itself. The two proposed applications are intended to illustrate use cases in which applying this idea is both convenient and advantageous.

\section{Acknowledgements}
\ifinterspeechfinal
    This work was supported by MCIN/AEI/10.13039/501100011033 under Grant PID2021-126061OB-C44, and in part by the Government of Aragón (Grant Group T36 23R). This project has received funding from the European Union’s Horizon 2020 research and innovation programme under the Marie Skłodowska-Curie grant agreement No 101007666.
\else
     Lorem ipsum dolor sit amet, consectetuer adipiscing elit. Aenean commodo ligula eget dolor. Aenean massa. Cum sociis natoque penatibus et magnis dis parturient montes, nascetur ridiculus mus. Donec quam felis, ultricies nec, pellentesque eu, pretium quis, sem. Nulla consequat massa quis enim. Donec pede justo, fringilla vel, aliquet nec, vulputa.
\fi

\bibliographystyle{IEEEtran}
\bibliography{mybib}

% Generated by IEEEtran.bst, version: 1.13 (2008/09/30)
\begin{thebibliography}{10}
\providecommand{\url}[1]{#1}
\csname url@samestyle\endcsname
\providecommand{\newblock}{\relax}
\providecommand{\bibinfo}[2]{#2}
\providecommand{\BIBentrySTDinterwordspacing}{\spaceskip=0pt\relax}
\providecommand{\BIBentryALTinterwordstretchfactor}{4}
\providecommand{\BIBentryALTinterwordspacing}{\spaceskip=\fontdimen2\font plus
\BIBentryALTinterwordstretchfactor\fontdimen3\font minus \fontdimen4\font\relax}
\providecommand{\BIBforeignlanguage}[2]{{%
\expandafter\ifx\csname l@#1\endcsname\relax
\typeout{** WARNING: IEEEtran.bst: No hyphenation pattern has been}%
\typeout{** loaded for the language `#1'. Using the pattern for}%
\typeout{** the default language instead.}%
\else
\language=\csname l@#1\endcsname
\fi
#2}}
\providecommand{\BIBdecl}{\relax}
\BIBdecl

\bibitem{rosch1973natural}
E.~H. Rosch, ``Natural categories,'' \emph{Cognitive psychology}, vol.~4, no.~3, pp. 328--350, 1973.

\bibitem{bengio2013representation}
Y.~Bengio, A.~Courville, and P.~Vincent, ``Representation learning: A review and new perspectives,'' \emph{IEEE transactions on pattern analysis and machine intelligence}, vol.~35, no.~8, pp. 1798--1828, 2013.

\bibitem{achille2018emergence}
A.~Achille and S.~Soatto, ``Emergence of invariance and disentanglement in deep representations,'' \emph{The Journal of Machine Learning Research}, vol.~19, no.~1, pp. 1947--1980, 2018.

\bibitem{klys2018learning}
J.~Klys, J.~Snell, and R.~Zemel, ``Learning latent subspaces in variational autoencoders,'' \emph{Advances in neural information processing systems}, vol.~31, 2018.

\bibitem{wen2016discriminative}
Y.~Wen, K.~Zhang, Z.~Li, and Y.~Qiao, ``A discriminative feature learning approach for deep face recognition,'' in \emph{Computer Vision--ECCV 2016: 14th European Conference, Amsterdam, The Netherlands, October 11--14, 2016, Proceedings, Part VII 14}.\hskip 1em plus 0.5em minus 0.4em\relax Springer, 2016, pp. 499--515.

\bibitem{lin2017focal}
T.-Y. Lin, P.~Goyal, R.~Girshick, K.~He, and P.~Doll{\'a}r, ``Focal loss for dense object detection,'' in \emph{Proceedings of the IEEE international conference on computer vision}, 2017, pp. 2980--2988.

\bibitem{ranasinghe2021orthogonal}
K.~Ranasinghe, M.~Naseer, M.~Hayat, S.~Khan, and F.~S. Khan, ``Orthogonal projection loss,'' in \emph{Proceedings of the IEEE/CVF international conference on computer vision}, 2021, pp. 12\,333--12\,343.

\bibitem{higgins2017beta}
I.~Higgins, L.~Matthey, A.~Pal, C.~P. Burgess, X.~Glorot, M.~M. Botvinick, S.~Mohamed, and A.~Lerchner, ``beta-vae: Learning basic visual concepts with a constrained variational framework.'' \emph{ICLR (Poster)}, vol.~3, 2017.

\bibitem{zhu2021and}
X.~Zhu, C.~Xu, and D.~Tao, ``Where and what? examining interpretable disentangled representations,'' in \emph{Proceedings of the IEEE/CVF Conference on Computer Vision and Pattern Recognition}, 2021, pp. 5861--5870.

\bibitem{pan2019transferrable}
Y.~Pan, T.~Yao, Y.~Li, Y.~Wang, C.-W. Ngo, and T.~Mei, ``Transferrable prototypical networks for unsupervised domain adaptation,'' in \emph{Proceedings of the IEEE/CVF conference on computer vision and pattern recognition}, 2019, pp. 2239--2247.

\bibitem{zinemanas2021interpretable}
P.~Zinemanas, M.~Rocamora, M.~Miron, F.~Font, and X.~Serra, ``An interpretable deep learning model for automatic sound classification,'' \emph{Electronics}, vol.~10, no.~7, p. 850, 2021.

\bibitem{mariotte2024explainable}
T.~Mariotte, A.~Almudévar, M.~Tahon, and A.~Ortega, ``An explainable proxy model for multiabel audio segmentation,'' 2024.

\bibitem{almudevar23_interspeech}
A.~Almudévar, A.~Ortega, L.~Vicente, A.~Miguel, and E.~Lleida, ``{Variational Classifier for Unsupervised Anomalous Sound Detection under Domain Generalization},'' in \emph{Proc. INTERSPEECH 2023}, 2023, pp. 2823--2827.

\bibitem{higgins2018towards}
I.~Higgins, D.~Amos, D.~Pfau, S.~Racaniere, L.~Matthey, D.~Rezende, and A.~Lerchner, ``Towards a definition of disentangled representations,'' \emph{arXiv preprint arXiv:1812.02230}, 2018.

\bibitem{locatello2019challenging}
F.~Locatello, S.~Bauer, M.~Lucic, G.~Raetsch, S.~Gelly, B.~Sch{\"o}lkopf, and O.~Bachem, ``Challenging common assumptions in the unsupervised learning of disentangled representations,'' in \emph{international conference on machine learning}.\hskip 1em plus 0.5em minus 0.4em\relax PMLR, 2019, pp. 4114--4124.

\bibitem{locatello2020weakly}
F.~Locatello, B.~Poole, G.~R{\"a}tsch, B.~Sch{\"o}lkopf, O.~Bachem, and M.~Tschannen, ``Weakly-supervised disentanglement without compromises,'' in \emph{International conference on machine learning}.\hskip 1em plus 0.5em minus 0.4em\relax PMLR, 2020, pp. 6348--6359.

\bibitem{estermann2023dava}
B.~Estermann and R.~Wattenhofer, ``Dava: Disentangling adversarial variational autoencoder,'' \emph{arXiv preprint arXiv:2303.01384}, 2023.

\bibitem{almudévar2024unsupervised}
A.~Almudévar, T.~Mariotte, A.~Ortega, and M.~Tahon, ``Unsupervised multiple domain translation through controlled disentanglement in variational autoencoder,'' 2024.

\bibitem{zhang2017mixup}
H.~Zhang, M.~Cisse, Y.~N. Dauphin, and D.~Lopez-Paz, ``mixup: Beyond empirical risk minimization,'' \emph{arXiv preprint arXiv:1710.09412}, 2017.

\bibitem{johnson1986extensions}
W.~B. Johnson, J.~Lindenstrauss, and G.~Schechtman, ``Extensions of lipschitz maps into banach spaces,'' \emph{Israel Journal of Mathematics}, vol.~54, no.~2, pp. 129--138, 1986.

\bibitem{ailon2006approximate}
N.~Ailon and B.~Chazelle, ``Approximate nearest neighbors and the fast johnson-lindenstrauss transform,'' in \emph{Proceedings of the thirty-eighth annual ACM symposium on Theory of computing}, 2006, pp. 557--563.

\bibitem{kingma2013auto}
D.~P. Kingma and M.~Welling, ``Auto-encoding variational bayes,'' \emph{arXiv preprint arXiv:1312.6114}, 2013.

\bibitem{piczak2015esc}
K.~J. Piczak, ``Esc: Dataset for environmental sound classification,'' in \emph{Proceedings of the 23rd ACM international conference on Multimedia}, 2015, pp. 1015--1018.

\bibitem{warden2018speech}
P.~Warden, ``Speech commands: A dataset for limited-vocabulary speech recognition,'' \emph{arXiv preprint arXiv:1804.03209}, 2018.

\bibitem{busso2008iemocap}
C.~Busso, M.~Bulut, C.-C. Lee, A.~Kazemzadeh, E.~Mower, S.~Kim, J.~N. Chang, S.~Lee, and S.~S. Narayanan, ``Iemocap: Interactive emotional dyadic motion capture database,'' \emph{Language resources and evaluation}, vol.~42, pp. 335--359, 2008.

\bibitem{yang2021superb}
S.-w. Yang, P.-H. Chi, Y.-S. Chuang, C.-I.~J. Lai, K.~Lakhotia, Y.~Y. Lin, A.~T. Liu, J.~Shi, X.~Chang, G.-T. Lin \emph{et~al.}, ``Superb: Speech processing universal performance benchmark,'' \emph{arXiv preprint arXiv:2105.01051}, 2021.

\bibitem{desplanques2020ecapa}
B.~Desplanques, J.~Thienpondt, and K.~Demuynck, ``Ecapa-tdnn: Emphasized channel attention, propagation and aggregation in tdnn based speaker verification,'' \emph{arXiv preprint arXiv:2005.07143}, 2020.

\bibitem{gong2021ast}
Y.~Gong, Y.-A. Chung, and J.~Glass, ``Ast: Audio spectrogram transformer,'' \emph{arXiv preprint arXiv:2104.01778}, 2021.

\bibitem{deng2009imagenet}
J.~Deng, W.~Dong, R.~Socher, L.-J. Li, K.~Li, and L.~Fei-Fei, ``Imagenet: A large-scale hierarchical image database,'' in \emph{2009 IEEE conference on computer vision and pattern recognition}.\hskip 1em plus 0.5em minus 0.4em\relax Ieee, 2009, pp. 248--255.

\bibitem{gemmeke2017audio}
J.~F. Gemmeke, D.~P. Ellis, D.~Freedman, A.~Jansen, W.~Lawrence, R.~C. Moore, M.~Plakal, and M.~Ritter, ``Audio set: An ontology and human-labeled dataset for audio events,'' in \emph{2017 IEEE international conference on acoustics, speech and signal processing (ICASSP)}.\hskip 1em plus 0.5em minus 0.4em\relax IEEE, 2017, pp. 776--780.

\bibitem{chen2022beats}
S.~Chen, Y.~Wu, C.~Wang, S.~Liu, D.~Tompkins, Z.~Chen, and F.~Wei, ``Beats: Audio pre-training with acoustic tokenizers,'' \emph{arXiv preprint arXiv:2212.09058}, 2022.

\bibitem{murray1993toward}
I.~R. Murray and J.~L. Arnott, ``Toward the simulation of emotion in synthetic speech: A review of the literature on human vocal emotion,'' \emph{The Journal of the Acoustical Society of America}, vol.~93, no.~2, pp. 1097--1108, 1993.

\bibitem{hammerschmidt2007acoustical}
K.~Hammerschmidt and U.~J{\"u}rgens, ``Acoustical correlates of affective prosody,'' \emph{Journal of voice}, vol.~21, no.~5, pp. 531--540, 2007.

\bibitem{sauter2010perceptual}
D.~A. Sauter, F.~Eisner, A.~J. Calder, and S.~K. Scott, ``Perceptual cues in nonverbal vocal expressions of emotion,'' \emph{Quarterly journal of experimental psychology}, vol.~63, no.~11, pp. 2251--2272, 2010.

\bibitem{talkin1995robust}
D.~Talkin and W.~B. Kleijn, ``A robust algorithm for pitch tracking (rapt),'' \emph{Speech coding and synthesis}, vol. 495, p. 518, 1995.

\bibitem{recommendation2015itu}
I.~Recommendation, ``Itu-r bs. 1770-4,'' \emph{Algorithms to measure audio programme loudness and true-peak audio level}, 2015.

\end{thebibliography}

\end{document}